# MLPrE—A tool for preprocessing and exploratory data analysis prior to machine learning model construction

David S Maxwell[1,2], Michael Darkoh[2], Sidharth R Samudrala[2], Caroline Chung[1,3], Stephanie T Schmidt[2,3], Bissan Al-Lazikani[2,3]

[1] Data Impact and Governance, The University of Texas MD Anderson Cancer Center, Houston, Texas, USA
[2] Department of Genomic Medicine, The University of Texas MD Anderson Cancer Center, Houston, Texas, USA
[3] The Institute for Data Science in Oncology, The University of Texas MD Anderson Cancer Center, Houston, Texas, USA

Corresponding Author:
David S Maxwell[1]
South Campus Research Building, 4SCRB1.1048, Houston, Texas, 77030 USA
Email address: dmaxwell@mdanderson.org

# Abstract

**Background.** With the recent growth of Deep Learning for AI, there is a need for tools to meet the demand of data flowing into those models. In some cases, source data may exist in multiple formats that negatively impact a constructed model. These formats may include varying states of preparation, errors and/or missing data, and therefore the source data must be investigated and properly engineered to match the needs for a Machine Learning (ML) model or graph database. Overhead and lack of scalability with existing workflows limit integration within a larger processing pipeline such as Apache Airflow, driving the need for a robust, extensible, and lightweight tool to preprocess arbitrary datasets and scale with the data type and size.

**Methods.** Herein, we describe a tool named Machine Learning Preprocessing and Exploratory Data Analysis or MLPrE. SparkDataFrames were utilized to hold data during processing and ensure scalability. A generalizable JSON input file format was utilized to describe stepwise changes to that DataFrame. Stages were implemented for input/output, filtering, basic statistics, feature engineering, and exploratory data analysis. Proof-of-concept for various MLPrE stages was demonstrated utilizing the following datasets: (1) Wilt sensing (2) bioconcentration, (3) myocardial infarction complications, (4) occupancy detection, (5) UniProt glossary data, and (6) phosphosite kinase datasets.

**Results.** A total of 69 stages were implemented into MLPrE. We highlighted key stages and demonstrated their utility throughout: the feature enrichment stage was tested using the Wilt dataset, the addMathExpression stage type tested using a QSAR bioconcentration dataset, exploratory data analysis stages tested using myocardial infarction datasets, and plotting stages tested with the occupancy datasets. We further highlight MLPrE's ability to independently process multiple fields in flat files and recombine them—which would otherwise require an additional pipeline—using a UniProt glossary term dataset. Building on this advantage, we demonstrated the clustering stage with available wine quality data which included two files: one for red wine and one for white wine. Lastly, we demonstrate the preparation of data for a graph database in the final stages of MLPrE using phosphosite kinase data. Overall, our MLPrE tool offers a generalizable and scalable tool for preprocessing and early data analysis, filling a critical need for such a tool given the ever-expanding use of machine learning. This tool serves to accelerate and simplify early-stage development in larger workflows.

# Introduction

Data preparation and cleaning consumes a significant percentage of time spent by a Data Scientist, yet it is critical to be done reliably and accurately. In fact, a 2020 survey by Anaconda shows that 45% of work time is spent doing those tasks, split into 19% for data loading and 26% in data cleansing (Anaconda 2020). While Data Engineers have largely taken over development for production level work, early-stage work for a project still necessitates data preprocessing. This is especially true when the project has not been fully developed, nor is it clear which of the parts of data will be needed for model construction.

Furthermore, there is an associated need to more comprehensively understand the data, necessitating rigorous exploratory data analysis (EDA) by the Data Scientist.

Data science notebooks such as Jupyter (Kluyver et al. 2016), Apache Zeppelin (Apache Zeppelin Team 2021), and Databricks notebook (Databricks 2022) are commonly utilized for early stage development work. Each notebook offers cells to place code (i.e., markdown) with options for running either individual or multiple cells in succession. They offer flexibility when trialing ideas during the crucial development phase. When moving to the production phase, though, these same features lead to other issues such as maintainability and reproducibility. Additionally, scaling to the full dataset during production often leads to memory or performance issues not observed during testing on data subsets during testing. Moving code developed in the notebook into a script is a typical solution but does not always address the above issues.

A common source of memory issues is processing DataFrames (Bohorquez 2021) using Pandas Python library (The pandas development team 2020). DataFrames are a common abstract for storing tabular style data and are routinely utilized in Data Science due to their resemblance to database tables. They are also implemented in the R statistical Package (R Core Team 2021). More recently, Spark DataFrames (Armbrust et al. 2015) have quickly become a standard for storing and manipulating large amounts of tabular data due to its lack of memory issues. Furthermore, because they were built upon the distributed nature of Apache Spark (Zaharia et al. 2016), they were made for scaling to larger datasets that may be encountered during production level activities. Pipeline and workflows both describe the process of filtering and transforming data by connecting components together and accomplishing a much larger task. Pipelines may be focused on the larger orchestration or lower-level data transformations and may be created visually through a script or driven by a data interchange format, such as JSON or YAML. On the visual side and lower level, KNIME (Berthold et al. 2008; Berthold et al. 2009) is a graphical ETL/BI tool that offers an intuitive drag-and-drop approach to data transformations. KNIME, however, presents drawbacks such as overhead and lack of scalability without moving from desktop to the server versions, which involves fees. A similar commercial product in this category of graphical tools for workflow creation is Alteryx Designer (Alteryx 2022). Orange is an open-source graphical tool that has been applied to single-cell data but it appears to be limited on the input types (Strazar et al. 2019).

Unix/Linux scripts have been utilized to address the needs of automating a series of operations. They offer a high degree of customization, standard syntax for file operations, and generally involve calling curated code that was developed over many years. Typically, the code is very specific and requires significant knowledge to modify. These types of scripts are found in all areas of science: bioinformatics (Software Carpentry Foundation 2016), cheminformatic s(Lee et al. 2017). Additional work in computational biology on a Script of Scripts (SoS) has addressed some of the learning curve and pain points associated with running those smaller scripts (Wang & Peng 2019). Apache Airflow (Apache Airflow Team 2021) is open-source option for connecting those individual scripts and allowing for concurrent running across a multiplatform environment. Regardless, there is still a need to standardize on a set of manipulations for preprocessing data for an automated pipeline involving the consensus of multiple ML methods, which can be scheduled to run on a regular basis. We believe focus on the smaller-data engineering transformations, as presented herein with MLPrE, complements existing software like Airflow and avoids replication of these types of higher-level workflow operations like scheduling.

JavaScript Object Notation (JSON) is a common format to exchange data on the web. It is easy for humans to read JSON and for machines to parse and generate. This same format may be utilized to direct code and drive data workflows. For example, the "Pipelines as Code" feature in CA Enterprise continuous delivery director (Broadcom 2022) effectively utilizes JSON to represent workflows. The AWS Data Pipeline (Services 2022) also uses JSON for the pipeline definition file. A superset of JSON called YAML Ain't Markup Language (YAML) is even easier to read and edit. The structure of YAML is handled through indentation and it has some additional functionality. The machine learning CLI (v1) for Azure allows the definition of pipeline using YAML (Microsoft 2022). For these reasons, we have selected JSON for our pipeline tool.

In this paper we present a tool that aids in data preparation and analysis while also allowing for improvements of consistency and operational efficiency in downstream pipelines. Later, we adapted to preprocessing of multiple datasets for input into a graph database. Our MLPrE tool is able to scale from a laptop to a cluster and analyze varying sizes of data, while still being relatively lightweight. Additionally, we show it is designed to perform efficiently in a lengthy workflow and provide basic error handling. Support for EDA, plotting, and clustering are also implemented, supporting early data discovery. Our work provides a framework for better and more automated data preprocessing and can serve as a valuable tool to drug discovers and, more generally, data scientists.

## Materials & Methods
### Computing

Development and testing for MLPrE was conducted on an Apache Hadoop Cluster and a development server, both of which were deployed at MD Anderson's data center. The 16-node Hadoop cluster (six master nodes and ten worker nodes) has 600 cores, 6TB of memory, and runs Hortonworks Data Platform (HDP) 2.6.5. The development server is a virtual server with 72 cores, 245MB of memory, and runs RHEL 8.9. MLPrE was initially written in Python 3.6.5 and used v2.3.0 of Spark as distributed in the HDP environment but has since been migrated to Python 3.9.x and v3.1.3 of Spark. Apache Yarn (Apache Yarn Team 2021) was utilized as the resource manager for all job submissions on the Hadoop cluster. Distributed files were stored in HDFS and Apache Hive (Apache Hive Team 2021) was utilized as one of the data input/output options, otherwise files were stored locally. Some limited testing was also done on a PC Laptop running Windows 11 with WSL2 installed.

### Datasets

The following datasets were processed with MLPrE: (1) Wilt data set (Johnson et al. 2013) was downloaded from UCI Machine Learning Repository (available as https://archive.ics.uci.edu/ml/datasets/Wilt), (2) QSAR Bioconcentration data set (Grisoni et al. 2016; Grisoni et al. 2015) was downloaded from UCI Machine Learning Repository (available as https://archive.ics.uci.edu/ml/datasets/QSAR+Bioconcentration+classes+dataset), (3) Myocardial Infarction Complications data set was downloaded from UCI Machine Learning Repository (available as https://archive.ics.uci.edu/ml/datasets/Myocardial+infarction+complications), (4) Occupancy Detection data set was downloaded from UCI Machine Learning Repository (available as https://archive.ics.uci.edu/ml/datasets/Occupancy+Detection+), (5) Glossary data was downloaded from UniProt (The UniProt Consortium 2022) (https://www.uniprot.org/keywords?query=*), (6) Phosphosite Kinase data (Hornbeck et al. 2014) was obtained from https://www.phosphosite.org/staticDownloads.

### General JSON Details

MLPrE utilizes a concept of data transformations directed through a JSON formatted input file that includes "stages" as a key in the key/value pair. The value associated with this key are wrapped in brackets to ensure that each stage is processed in order. Data is passed through each stage as a Spark DataFrame. At minimum, each stage has a stageName and stageType. The stageName is only relevant to printing in log file, but the stageType is used to select the type of processing. Many stages also have a stageParam, which is specific to the type of stage and typically utilizes default values, if not specified. Stages may be done in any order that makes sense, but the first stage must be an input stage to generate the initial Spark DataFrame. The stages utlzed by MLPrE are summarized in **Fig. 1**.

### Basic Stages

The following basic stages have been implemented in MLPrE: Input, Output, Remove Null Records, Keep/Remove Filters, Show DataFrame, Sample DataFrame, Add literal values and Unique ID columns, Structured Query Language (SQL).

Input and output into the preprocessing code includes csv but are significantly oriented towards formats which support big data: orc, parquet, and hive. Additionally, there is a mechanism for using SQL statements to pull in the initial dataset from hive rather than reading an entire table. Typically, the input is done as the first stage, but since it is not restricted to just that stage, reading the same or a new file may be done after an output or save DataFrame stage. Output may occur at any stage to a csv using Spark or Pandas (involves a conversion to Pandas DataFrame). Depending on the number of records and columns, one of these methods will be preferable. Output to Hive is also possible as are the Hadoop style outputs to parquet and orc formats.

Flat files such as csv and tsv formatted files have the same columns of data throughout; however, authors may utilize those fields differently to combine similar types of information into one file. In the case of UniProt glossary (The UniProt Consortium 2022), two levels of hierarchy exist in the same file. Most of the records are sub-terms as specified in the name column and parent terms in the category column. The ten parent term records have nothing specified for the Category. The file could be split into two files, but that would add an additional file in the pipeline. To avoid this, a tool is needed to process the two parts separately and recombine them; we implement this utility into the stage called unionDataframes.

To support data cleaning, we have implemented the ability to drop records containing NULLs and filtering records in or out based on certain criteria. Filtering out supports the ability to remove records matching a list of values in a column, though it can be inverted to keep those matching records. Additionally, there is a filter stage with additional options to keep only specific records. Matching values in a column can be specified through a list, simple expression, or SQL type like pattern.

It is important to view results of processing of stages, especially during testing. Our MLPrE does through a DataFrame show (i.e., a wrapper around the existing Spark .show() function). Similarly, we implement the same utility for the DataFrame sampling stage, which is utilized to obtain a subset of the data.

There are times when you need a column containing all the same values which can be achieved through a stage that adds literal values. Similarly, it may be useful to have a unique id column; this has been implemented using the monotonically_increasing_id pyspark.sql.function, with the added ability to place the id column as first column in DataFrame. Due to the distributed nature of Spark, these numbers are not necessarily consecutive but are guaranteed to be increasing and unique, as documented in the Spark API reference.

In addition to allowing SQL at the input stage, we have implemented a stage that allows a SQL statement to query the existing state of the DataFrame. This is accomplished via creation of a temporary view in the code with the name "MyTempView" and using that as the table name in the SQL statement in stage parameters.

**Feature Engineering Stages**

The follow feature engineering stages have been implemented in MLPrE: Combine Columns, Math Operation and Expressions, String Replacement, DateTime Functions, Levenshtein Distance, Zscores, Min/Max Scaling, Encode Ranges. We describe each of these in detail subsequently.

When there is a need to merge multiple columns into one column, we added the combineColumns stage, which accomplishes this based on either a list or range of columns. There are options for a separator and removal of the existing columns. This stage might be used when creating a fingerprint based on single binary 1/0 values in multiple columns.

Due to the importance of math operations in feature engineering, several stage types have been implemented. Firstly, there is a simplistic math operation stage, addMathExpression, for performing expressions on multiple columns. As currently implemented, the expression is not parsed and will be evaluated in the order found in list. Thus, an expression such as col1 + col2 * col3 should be written as ["col2", "*", "col3", "+", "col1"] to get a proper result. Implementing a proper math parser would be preferred which we will consider in a later version. Secondly, there is a stage called addColumnMath, to do simple math functions (e.g., cos(x)) on a list of columns. Each column is verified as numeric before attempting the function. Also, there are options for using a prefix for naming new columns and specifying the number of decimal places. Lastly, there is a stage called applyColumnMath that runs simple operators (+, -, *, /) between a given list of columns and a single specified column. The selection of columns may be done through an include, exclude, or startswith, ensuring that the proper set of columns can be specified even with large lists of columns.

A basic replacement of strings in a text column was done using a stage type called replaceStrings. It goes through a list of specified columns and has a replacement dictionary with key as the search string and value as the replacement. It does a simple replace using the Spark regexg_replace function.

Date and time functions are another common area utilized in feature enrichment. There is a stage type, enrichDateTime, that applies one of the following datetime functions ("dayofweek", "dayofmonth", "dayofyear", "hour", "minute", "second", "month", "year", "weekofyear") to a list of columns, with optional specification of a suffix to apply to new columns created.

The Levenshtein function (Levenshtein 1966) is a known method for comparing two strings, for which a PySpark SQL function already exists. MLPrE's stage type, addLevenshtein, utilizes this function to compute the distance for all pair combinations of valid string columns from a given list.

Z-score normalization (Grafen & Hails 2002) and Min/Max scaling were implemented through the stage types, addZscores and addMinMaxScalings, respectively. Both stage types will compute the normalization based on given list of columns and then name the new columns with a specified suffix. There is also a single column version of each that will be replaced with a specified column name.

We implemented feature binning through a stage type called encodeRanges. The encoded column, ranges, and new column are specified in stageParam. "Minimum" and "Maximum" are specific keys that trigger the comparison for low end and high end of ranges, respectively. Anything less than or equal to Minimum picks up on the value corresponding to "Minimum" key, whereas anything greater than "Maximum" is set to the value of "Maximum". Other key names do not matter but should be unique.

## Plotting

The Seaborn library (Waskom 2021) was utilized to add plotting functionality to MLPrE. We focused on the ability to automate quick visualization of a larger number of plots versus wrapping each possible argument parameter. In the case of Boxplot, we implemented this for a combination of categorical and numeric columns, with categorical values shown on x-axis. Categorical values were specified in a list or determined automatically based on a cutoff when requesting all categorical columns. All possible numeric columns may be selected or provided in list form. Setting the processAllNumeric and processAllCategorical to true will override any values in the lists. This same strategy was utilized in coding for pair plots, histogram, and scatterplots.

## Clustering

Because clustering is an important tool in early data analysis, we have included a simple clustering stage (simpleCluster) that has a minimal set of options that allow for scanning the number of clusters using the KMeans or BisectingKMeans algorithms as implemented in Spark. A full review of clustering using Apache Spark, including the benefits for usage on Big Data, may be found elsewhere (Saeed et al. 2020). The clustering stage requires a prior stage called addFeaturesVector, which takes a list of columns and turns it into a vector column, with support for both categorical and numerical features. There is an optional parameter to apply MinMaxScaling to the vector. With a range of $k_{min}$ to $k_{max}$, a file containing the cluster costs are written out. As described in the Spark documentation, the costs are the "sum of squared distances of points to their nearest center". A plot of this data can lead to a selection of ideal cluster number using the graphical "elbow" method. Setting the $k_{min}$ equal to $k_{max}$ and running the clustering again will generate the predicted clusters.

The "Silhouette" method (Rousseeuw 1987) is available to directly evaluate the Spark clustering results and is commonly utilized in finding the best number of k clusters. The Silhouette scores range from -1 to 1, where a higher positive number is reflective of finding points closer with the same cluster and further away from other clusters. We have also implemented the algorithm described by Shi et al. (Shi et al. 2021) for identifying the optimal number of clusters. This uses an alternative approach involving angles, with a maximal value set at pi since the angle comes from an arccos function. The lowest angle value is the $K_{opt}$ value.

## Exploratory Data Analysis Stages

All prior stages lay the ground for MLPrE to support early-stage EDA work by Data Scientists. For example, the edaFeatureExtents stage type analyzes numeric columns and outputs min, max, mean, standard deviation, kurtosis, and skewness. The edaFeatureTypes stageTypes addresses the need to identify the total and distinct count per column, column type, and percentage filled, whether a column is all integer or if it is a categorical column based upon a cutoff. The edaCompleteObservations stage type performs a complete observations analysis on the dataframe. In particular, the percentage of nulls and blanks are noted for each column. Correlations between numeric columns are implemented through stage types edaColumnCorrelation or edaPairwareCorrelation. The column correlation performs correlations of a list of columns to one column, whereas the pairwise version allows for all combinations between those in a list. These functionalities employed by data scientists enable accelerated early-stage EDA work, which we illustrate lateron.

## Results

Shown in **Figure 2** is an example of feature enrichment utilizing the encodeRanges stage type on the Wilt data set. In the first stage, there were three encoded ranges for the column GLCM_pan that map to the strings "Low", "Med", and "High" in columns GPR. Values <= 110.0 were mapped to "Low". Values >100.0 and <=150.0 were mapped to "Med". Values >150.0 were mapped to "High". In the second stage, the encode was done for column SD_pan. Resulting ranges were mapped to integers from 1-5 in column SPR. Only the "Minimum" and "Maximum" key names matter. An example of the addMathExpression stage type is shown in **Figure 3** for the QSAR bioconcentration data. Presently, there is no mathematical parsing of the expression, so in practice, the expression may need to be arranged to ensure it is properly calculated. Alternatively, the sql stage could be utilized to perform the math expression instead.

Next, we performed an exploratory data analysis for the Myocardial infarction complication data set as shown in **Figure 4a**. The feature types analysis (FTA) shows overall and distinct counts for this data set; there were 1692 listed ages in the file, but only 62 unique values existed and the column type was listed as integer. It was clear from the percentFilled column that there were some missing values. Running a feature extent analysis would allow for additional analysis such as minimum and maximum on the age, ensuring that the values are within an expected range. Columns with a high percentage of null (or blank) values can be justifiably removed as potential features, such as observed with the KFK_BLOOD column in **Figure 4b** with nearly 100% null values. This could also suggest an issue with the data pull, but we did not find evidence of this in our investigation.

Plotting stages as implemented in MLPrE are depicted in **Figure 5A** along with examples of plotting capabilities using an occupancy dataset as shown in **Figure 5B-5C**. The first example is a scatter plot which shows Light vs Temperature with coloring of the points by occupancy values (0=unoccupied, 1=occupied), as shown in **Figure 5B**. It was clear from this simple example that higher light values were indicative of area being occupied. Notably, original publication (Candanedo & Feldheim 2016) associated with this data set showed the top node for both CART models was light, suggesting it was the most important factor for occupancy status. Another important factor was $CO_2$. A boxplot of that factor vs occupancy is shown as the second example in **Fig. 5C**. In this case, only one of the plot combinations is shown although different from what one might expect given "Temperature" and "Humidity" as numeric columns. Although there were values shown in the numericColumnNames and categoricalColumnNames, they were ignored since both the processAllNumeric and processAllCategorical Boolean parameters were set to true.

To avoid situations where authors may utilize information from multiple fields in flat files requiring two files and thus an additional pipeline, we needed to be able to process the two parts differently and recombine them. In **Figure 6a**, we show this type of processing. The UniProt glossary was downloaded as a TSV file, which contained the parent terms and sub-terms together. Matching the format for glossary in OpenMetadata required the sub-term records processed with glossaryname.parent for parent column and parent term records with empty string. For brevity, we show only the final stages with the parent section already stored. In stage 19, a literal value was created, followed by a stage to combine with parent name and rename the column back to parent. A union stage combined the stored DataFrame with a modified sub-term DataFrame. Finally, a select was done to ensure correct column ordering before writing out as a CSV file. A section of the output is shown in **Figure 6b**. The result from loading the data into OpenMetadata is shown in **Figure 6c**.

We selected a wine quality dataset to test the simple clustering stage in MLPrE. The wine quality was contained in two files, one each for red and white wine. During the preparation phase, the two files were combined, spaces were removed from column names, and a unique_id column was added for later use. During the clustering phase, all numerical features were utilized, and a clustering was performed from $k_{min}=2$ to $k_{max}=20$ with costs written out. With the knowledge of seven different quality levels from the source data, a clustering prediction was generated with k=8 since a view of the costs versus k did not immediately lead to an obvious selection of optimal value. This was followed by generation of boxplots for all numeric features vs predicted cluster. In **Figure 7a**, we show total sulfur dioxide with a reasonable separation of values for each predicted cluster. The same is true for values of alcohol feature. Other plots such as chlorides and alcohol **(Fig. 7b)** showed much less separation. Insights from these boxplots and others provide a better of understanding of the data prior to building full classification models. We found that the optimal number of clusters via the "Silhouette" method to be 4 followed by 3 and 5. This compares well with the optimal value of 3 identified using Shi et al.'s method (Shi et al. 2021).

We have selected the Phosphosite Kinase data as an example of usage of MLPrE to prepare data for a graph database, in this case Neo4J. We started with a tab delimited file containing kinase and substrate information, for which we wanted a csv formatted file that could be easily processed by the Cypher "LOAD CSV" commands. In the first stage, the file was read as a tab delimited file and a subset of columns was selected. In our graph, we were only interested in the interactions for homo sapiens, so two filter stages were utilized to filter based on a list with a single value of "human". The next two stages handle obtaining the parent accession by obtaining the substring before the dash (e.g., P05771-2 to P05771), allowing for a connection through a parent node rather than through a specific isoform when loading the data. Finally, the dataset was written out.

## Discussion

A major advantage of MLPrE is that data input can come in the form of flat files, through querying of Hive tables, or by reading other common Hadoop file formats such as parquet. Additionally, we adopted Spark DataFrames as the mechanism to pass data between each of the steps, allowing it to scale and avoid potential memory or processing constraints on larger data sets. The JSON utilized for directing the underlying PySpark code is self-documenting and easily adaptable to new projects. The inclusion of new stages has largely been driven by our own processing needs, therefore it is possible that it does not cover all type of preprocessing scenarios but adding new stages could be easily implemented. The current version is limited to a linear series of stages, though features such as the ability to store and recall a DataFrame have reduced that impact. MLPrE would also benefit from a graphical user interface to assist with building processing templates and interactive debugging. More recent versions of Spark (>=3.4) have a decoupled client-server that would support building a GUI and we are considering that for a future release.

Despite these limitations, the utility of this code is the ability to run anywhere that Spark and Python can be installed. When incorporated into a much larger workflow such as Apache Airflow(Apache Airflow Team 2021), the tool becomes even more powerful due to not necessitating changes in the code by making minor changes in the way the data is processed. Our pipeline saves many hours of setup and debugging per run, since the only change was an input JSON. The stages are expressed in a user friendly way, facilitating fast changes to processing without exploration into existing code, which is ideal for collaborative work. MLPrE has also saved valuable time by running EDA for other projects and checking the output of processed data.

## Conclusions

We developed MLPrE as a piece of stand-alone code for preprocessing and early development analysis needs in Data Science and Data Engineering. Our tool builds on JSON file formats, which we implemented because its stage architecture was suitable for our preprocessing needs. Stages for input/output, filtering, SQL, and basic modifications to columns were implemented. In terms of feature engineering, we included several simple math and string operations, range encoding, and datetime functions. Recognizing that EDA is a critical component to data science and data engineering, stages were constructed for column correlations, checking feature extents, types, and determining complete observations. Additionally, several plotting stages with common configurable options were implemented, allowing for easy exploration of data. Processing flat files where similar data has been processed and combined into one file can present challenges for many processing pipelines, but we demonstrated MLPrE overcomes this challenge using an example from UniProt glossary data. Lastly, a stage for clustering analysis was implemented as another EDA stage.

Each of these stages was tested against unique datasets, highlighting distinct use cases for MLPrE. The chosen datasets show applicability to various biological and clinical settings, for example the QSAR bioconcentration data or myocardial infarction datasets, lending MLPrE as a suitable preprocessing and EDA tool for early-stage drug discovers. More generally, MLPrE serves as a tool offering basic, EDA, plotting, and feature engineering stages, providing needed support for data scientists and engineers during preprocessing. When utilized properly by larger workflows, the MLPrE serves to accelerate and simplify early-stage development. Currently, there are plans to expand the MLPrE through adding additional stages, global variables, and incorporating other types of DataFrames.

## Acknowledgements

This work was conceived and initially worked on while David Maxwell was a member of the DS team in the EDEA department at UTMDACC. We would like to thank Melody Page for managing the team in EDEA and Brian Dyrud for handling business system analysis work during the project execution. Additionally, we would also like to thank other members of the EDEA team: Jim Lomax, Robert Brown, and Ya Zhang for their valuable suggestions regarding additional stages for MLPrE. We would also like to acknowledge the efforts of Mary McGuire for reading through and commenting on early drafts of the manuscript. We would also like to express our appreciation to Daniel Wang for administration of the Hadoop Cluster, Apache Airflow, and for debugging issues that arose during development of MLPrE. This work was supported by The Institute for Data Science in Oncology (IDSO) and Context Engine within UTMDACC. B.A-L. is a Cancer Prevention & Research Institute of Texas (CPRIT) Scholar in Cancer Research and is grateful for their support. She is funded under the Cancer Prevention and Research Institute of Texas (CPRIT) Established Investigator Award (RR210007).

## Additional Information and Declarations
### Funding
Cancer Prevention and Research Institute of Texas (CPRIT) Established Investigator Award [RR210007], (BA-L)
The Commonwealth Foundation, (BA-L)
The Lyda Hill Foundation, (BA-L)
CRUK Drug Discovery Committee strategic award, (BA-L)
NIHR (National Institute of Health Research) support to the Biomedical Research Centre at the Institute of Cancer Research, (BA-L)
Royal Marsden NHS Foundation Trust, (BA-L)

### Grant Disclosures
Cancer Prevention and Research Institute of Texas (CPRIT) Established Investigator Award [RR210007] (BA-L)

### Competing Interests
B.A.-L. declares financial interest in Exscientia PLC, Recursion Pharmaceuticals Inc, Drug Hunter and AstraZeneca PLC. B.A.-L. is/has been a member of Scientific Advisory Boards and/or provided paid consultancy for the following: Astex Pharmaceuticals, AstraZeneca PLC, GSK PLC, Novo Nordisk, Sante Ventures. She is a lead on the MD Anderson Drug Discovery and Development Division which has commercial interest in target and drug discovery. She was a chair and member of the Scientific Advisory Board for Open Targets. She is chair of the Cancer Research UK Data Strategy Board and member of the CRUK Scientific Advisory Board. She is a member of the Board of Directors on the Leukemia and Lymphoma Society and a member of the New York Genome Consortium Scientific Advisory Board. She is Director of non-profit Chemical Probes Portal. DM, MD, SS, CC, STS, and BA-L are employees of UT MD Anderson Cancer Center which operates a reward to inventor scheme. Kaitlyn P. Russell served as the scientific writer of this paper and provided significant writing and revisions to the approved draft.

## Author Contributions

David Maxwell conceived the research, wrote, and tested the MLPrE code, prepared figures and/or tables, authored and reviewed drafts of the paper, and approved the final draft.

Michael Darkoh conducted testing and provided feedback on changes to the preprocessing code.

Sidarth Samudrala implemented and tested an algorithm to determine optimal selection of k clusters as part of the clustering stage.

Caroline Chung conceived of the initial project work that became the impetus for MLPrE, reviewed output from consensus ML models resulting from usage of the code as part of a larger automated workflow.

Stephanie T. Schmidt provided guidance on development of the several processing stages that were utilized for preprocessing biological data for input into Neo4J, reviewed drafts of the paper.

Bissan Al-Lazikani conceived of the project that involved preprocessing biological datasets for input into Neo4J, supported the reviewed drafts of the paper, approved the final draft.

## Data Availability

MLPrE's code and the data needed to replicate the work performed herein are supplied as additional supplementary files.

## References

Alteryx. 2022. Alteryx Designer. *Available at* https://www.alteryx.com/products/alteryx-designer.
Anaconda. 2020. The State of Data Science 2020 Moving from hype toward maturity. *Available at* https://www.anaconda.com/state-of-data-science-2020?utm_medium=press&utm_source=anaconda&utm_campaign=sods-2020&utm_content=report.
Apache Airflow Team. 2021. Apache Airflow Project. *Available at* https://airflow.apache.org/.
Apache Hive Team. 2021. Apache Hive Project. *Available at* https://hive.apache.org/.
Apache Yarn Team. 2021. Apache Yarn Project. *Available at* https://hadoop.apache.org/docs/stable/hadoop-yarn/hadoop-yarn-site/YARN.html.

Apache Zeppelin Team. 2021. Apache Zeppelin Project. *Available at https://zeppelin.apache.org/*.

Armbrust M, Xin RS, Lian C, Huai Y, Liu D, Bradley JK, Meng XR, Kaftan T, Franklint MJ, Ghodsi A, and Zaharia M. 2015. Spark SQL: Relational Data Processing in Spark. *Sigmod'15: Proceedings of the 2015 Acm Sigmod International Conference on Management of Data*:1383-1394. 10.1145/2723372.2742797

Berthold MR, Cebron N, Dill F, Gabriel TR, Kotter T, Meinl T, Ohl P, Sieb C, Thiel K, and Wiswedel B. 2008. KNIME: The Konstanz Information Miner. *Data Analysis, Machine Learning and Applications*:319-326. Doi 10.1145/1656274.1656280

Berthold MR, Cebron N, Dill F, Gabriel TR, Kötter T, Meinl T, Ohl P, Thiel K, and Wiswedel B. 2009. KNIME-the Konstanz information miner: version 2.0 and beyond. *AcM SIGKDD explorations Newsletter* 11:26-31.

Bohorquez N. 2021. How to avoid Memory errors with Pandas (accessed 4/23/2024 2024).

Broadcom. 2022. Continuous Delivery Director. *Available at https://techdocs.broadcom.com/us/en/ca-enterprise-software/devops/continuous-delivery-director-onprem/8-4/releases/release-design/pipelines-as-code.html*.

Candanedo LM, and Feldheim V. 2016. Accurate occupancy detection of an office room from light, temperature, humidity and $CO_2$ measurements using statistical learning models. *Energy and Buildings* 112:28-39. 10.1016/j.enbuild.2015.11.071

Databricks. 2022. Using Databricks Notebooks. *Available at https://docs.databricks.com/notebooks/*.

Grafen A, and Hails R. 2002. Modern statistics for the life sciences / Alan Grafen, Rosie Hails. Oxford ;: Oxford University Press, 319-320.

Grisoni F, Consonni V, Vighi M, Villa S, and Todeschini R. 2016. Investigating the mechanisms of bioconcentration through QSAR classification trees. *Environment International* 88:198-205. 10.1016/j.envint.2015.12.024

Grisoni F, Consonni V, Villa S, Vighi M, and Todeschini R. 2015. QSAR models for bioconcentration: Is the increase in the complexity justified by more accurate predictions? *Chemosphere* 127:171-179. 10.1016/j.chemosphere.2015.01.047

Hornbeck PV, Zhang B, Murray B, Kornhauser JM, Latham V, and Skrzypek E. 2014. PhosphoSitePlus, 2014: mutations, PTMs and recalibrations. *Nucleic Acids Research* 43:D512-D520. 10.1093/nar/gku1267

Johnson BA, Tateishi R, and Hoan NT. 2013. A hybrid pansharpening approach and multiscale object-based image analysis for mapping diseased pine and oak trees. *International Journal of Remote Sensing* 34:6969-6982. 10.1080/01431161.2013.810825

Kluyver T, Ragan-Kelley B, Perez F, Granger B, Bussonnier M, Frederic J, Kelley K, Hamrick J, Grout J, Corlay S, Ivanov P, Avila D, Abdalla S, Willing C, and Team JD. 2016. Jupyter Notebooks-a publishing format for reproducible computational workflows. *Positioning and Power in Academic Publishing: Players, Agents and Agendas*:87-90. 10.3233/978-1-61499-649-1-87

Lee ML, Aliagas I, Feng JWA, Gabriel T, O'Donnell TJ, Sellers BD, Wiswedel B, and Gobbi A. 2017. chemalot and chemalot_knime: Command line programs as workflow tools for drug discovery. *Journal of Cheminformatics* 9. ARTN 38

10.1186/s13321-017-0228-9

Levenshtein V. 1966. Binary codes capable of correcting deletions, insertions, and reversals. *Soviet Physics Doklady* 10:707-710.

Microsoft. 2022. Define Machine Learning Pipelines in YAML.

R Core Team. 2021. R: A Language and Environment for Statistical Computing. *Available at https://www.R-project.org/*.

Rousseeuw PJ. 1987. Silhouettes: A graphical aid to the interpretation and validation of cluster analysis. *Journal of Computational and Applied Mathematics* 20:53-65. https://doi.org/10.1016/0377-0427(87)90125-7

Saeed MM, Al Aghbari Z, and Alsharidah M. 2020. Big data clustering techniques based on Spark: a literature review. *PeerJ Comput Sci* 6:e321. 10.7717/peerj-cs.321

Services AW. 2022. AWS Data Pipeline - Developer Guide. *Available at* https://docs.aws.amazon.com/datapipeline/latest/DeveloperGuide/datapipeline-dg.pdf#what-is-datapipeline.

Shi C, Wei B, Wei S, Wang W, Liu H, and Liu J. 2021. A quantitative discriminant method of elbow point for the optimal number of clusters in clustering algorithm. *EURASIP Journal on Wireless Communications and Networking* 2021:31. 10.1186/s13638-021-01910-w

Software Carpentry Foundation. 2016. A simple bioinformatics workflow.

Strazar M, Zagar L, Kokosar J, Tanko V, Erjavec A, Policar PG, Staric A, Demsar J, Shaulsky G, Menon V, Lemire A, Parikh A, and Zupan B. 2019. scOrange-a tool for hands-on training of concepts from single-cell data analytics. *Bioinformatics* 35:i4-i12. 10.1093/bioinformatics/btz348

The pandas development team. 2020. pandas-dev/pandas: Pandas. 10.5281/zenodo.3509134

The UniProt Consortium. 2022. UniProt: the Universal Protein Knowledgebase in 2023. *Nucleic Acids Research* 51:D523-D531. 10.1093/nar/gkac1052

Wang G, and Peng B. 2019. Script of Scripts: A pragmatic workflow system for daily computational research. *PLoS computational biology* 15:e1006843-e1006843. 10.1371/journal.pcbi.1006843

Waskom ML. 2021. Seaborn: statistical data visualization. *Journal of Open Source Software* 6:3021. 10.21105/joss.03021

Zaharia M, Xin RS, Wendell P, Das T, Armbrust M, Dave A, Meng X, Rosen J, Venkataraman S, Franklin MJ, Ghodsi A, Gonzalez J, Shenker S, and Stoica I. 2016. Apache Spark: a unified engine for big data processing. *Commun ACM* 59:56–65. 10.1145/2934664

# Figures and Figure Legends

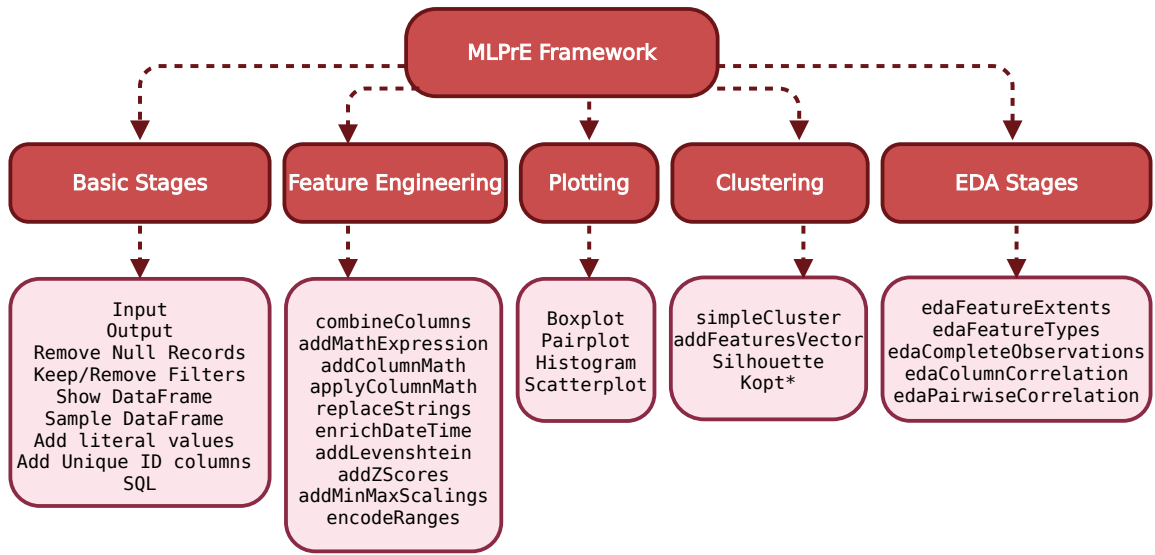

**Figure 1. Stages implemented into the MLPrE framework.** Stages are represented by red boxes, with associated functionalities in light red boxes.

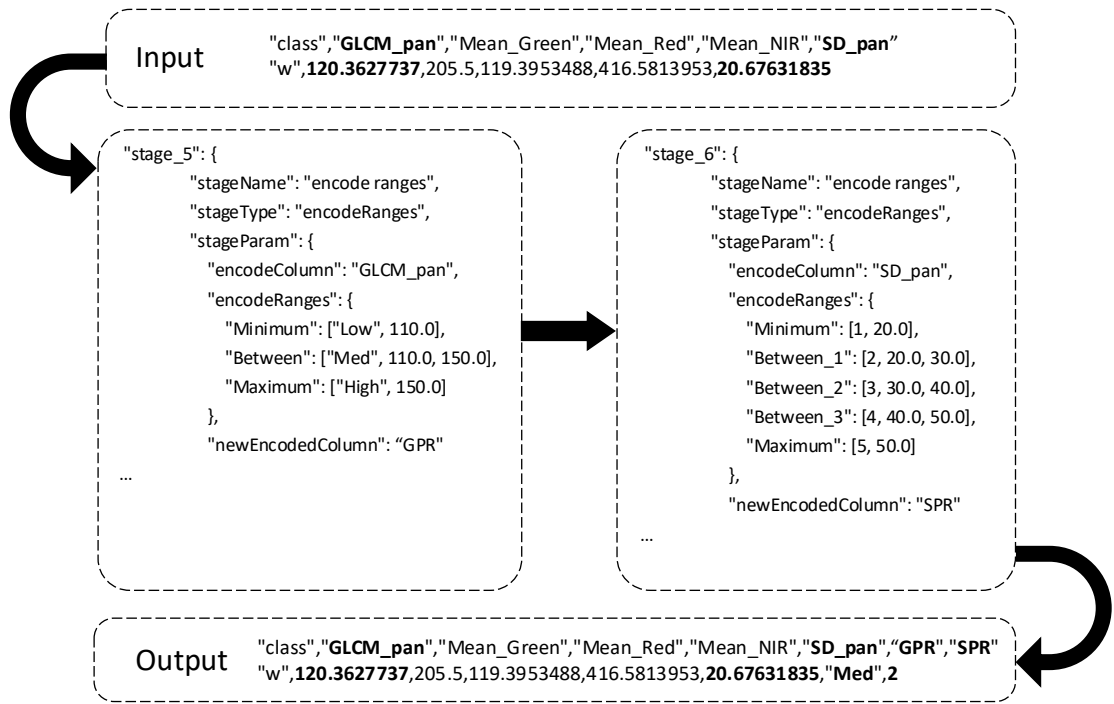

**Figure 2: Feature enrichment example usage of encodeRanges stage type using the Wilt data set.** Stages from the JSON input file show consecutive usage of the encodeRanges stage type to generate two new columns, GPR and SPR.

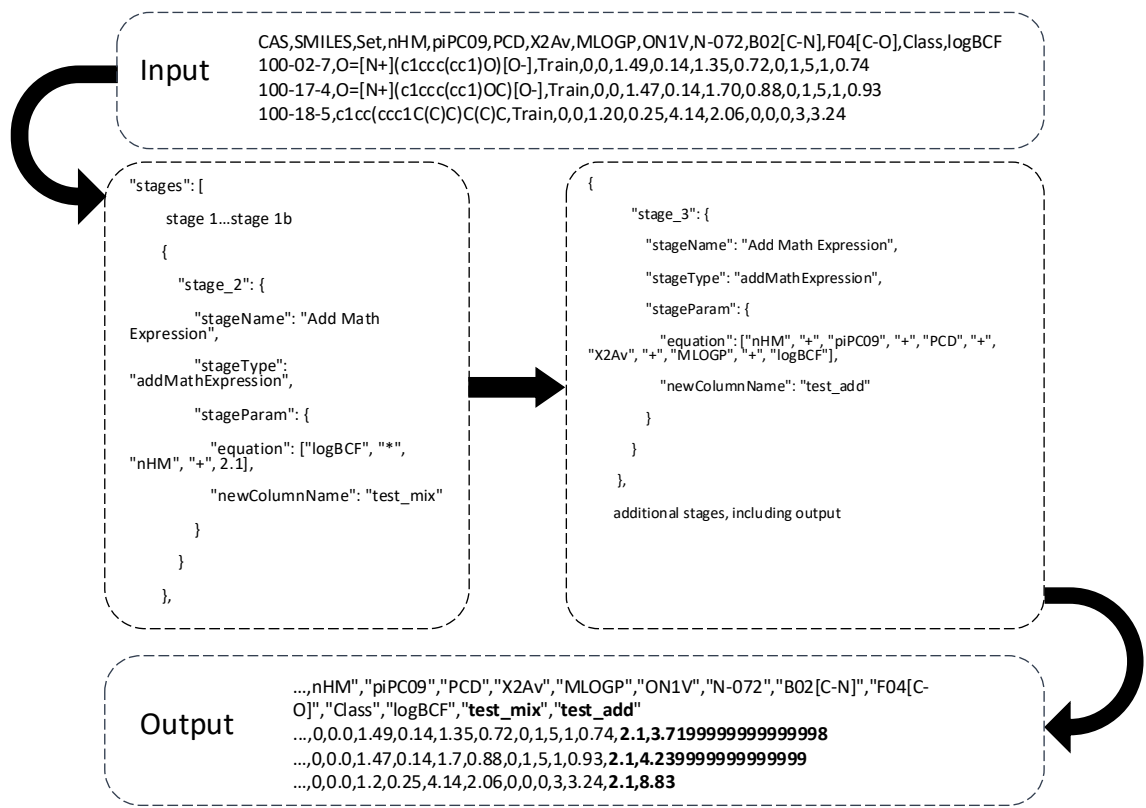

**Figure 3: Feature enrichment example usage of addMathExpression stage type using the QSAR Bioconcentration data set.** Stages from the JSON input file demonstrate consecutive usage of the addMathExpression stage type to generate two new columns. The nonsensical expressions demonstrate combinations of columns with some of the operators available in the stage type.

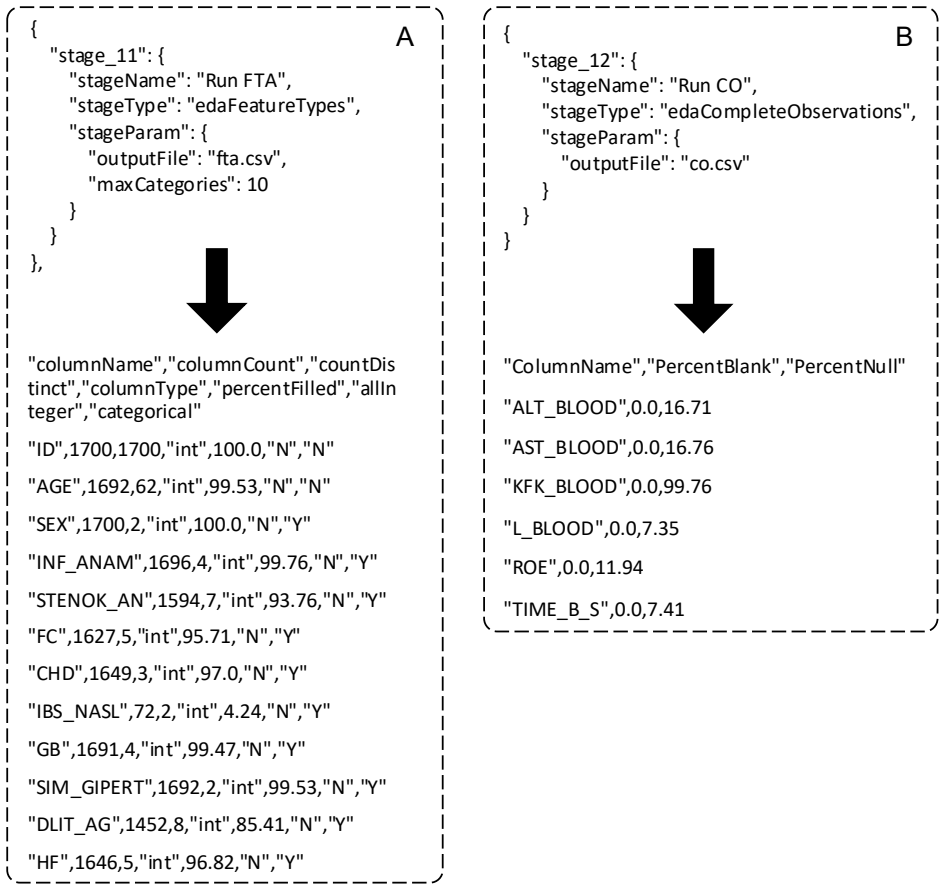

Figure 4: Exploratory Data Analysis (EDA) example of edaFeatureType and edaCompleteObservation stage types using the Myocardial Infarction Complication data set. Stages show two of the available EDA stage types. a) The stage edaFeaturesTypes gives basic information regarding the columns such as column types, counts, and percent filled. b) The stage edaCompleObservations shows information related to complete observations.

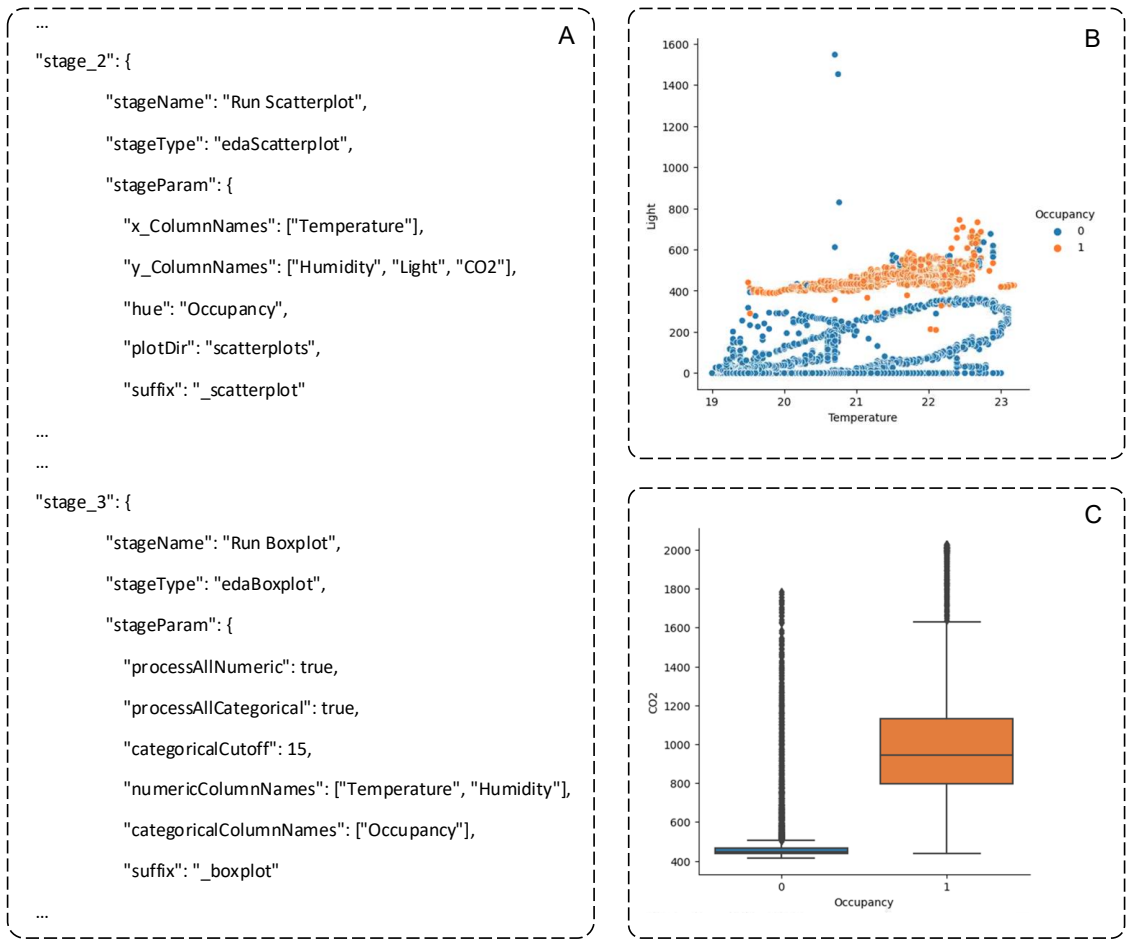

Figure 5: Plotting example of edaScatterplot and edaBoxplot stage types using the Occupancy Detection data set. (A) Code used to generate plots in (B) and (C). (B) One of three plots generated when using Temperature on x-axis and either Humidity, Light, or $CO_2$ selected for the y-axis. We have shown Light vs Temperature with coloring by the occupancy value. (C) All combinations of numeric and categorical columns, with categorical columns being determine based on a cutoff of 15. We selected the $CO_2$ vs Occupancy plot from these combinations.

```
{
    "stage_19": {
        "stageName": "Add Column",
        "stageType": "addLiteral",
        "stageParam": {
            "columnName": "TMP",
            "literalValue": "Interactome"
        }
    },
    "stage_20": {
        "stageName": "Combine columns",
        "stageType": "combineColumns",
        "stageParam": {
            "columnNames": ["TMP", "parent"],
            "newColumnName": "tmp_parent",
            "separator": ".",
            "removeColumns": true
        }
    },
    "stage_21": {
        "stageName": "Change column name",
        "stageType": "renameColumn",
        "stageParam": {
            "columnName": "tmp_parent",
            "newColumnName": "parent"
        }
    },
    "stage_22": {
        "stageName": "Union Dataframes",
        "stageType": "unionDataframes",
        "stageParam": {
            "unionType": "byName"
        }
    },
    "stage_23": {
        "stageName": "Select columns",
        "stageType": "selectColumns",
        "stageParam": {
            "columns": ["parent", "name*",
            "displayName", "description", "synonyms",
            "relatedTerms", "references", "tags", "reviewers",
            "owner", "status"]
        }
    },
    "stage_24": {
        "stageName": "Output",
        "stageType": "output",
        "stageParam": {
            "outputType": "csv",
            "outputFile": "sample_glossary_pre.csv",
            "header": true,
            "delimiter": ","
        }
    }
}
```

A

C

Interactome
Interactome

Terms   Activity Feeds & Tasks

Description
This is a glossary associated with Interactome.

Expand All

| TERMS | DESCRIPTION | OWNER | STATUS | ACTIONS |
|---|---|---|---|---|
| > Biological process | Keywords assigned to proteins because they are involved in a particular biological process. | No Owner | Approved | + ✎ |
| > Cellular component | Keywords assigned to proteins because they are found in a specific cellular or extracellular component. | No Owner | Approved | + ✎ |
| > Coding sequence diversity | Keywords assigned to proteins because their sequences can differ, due to differences in the coding sequences such as... more | No Owner | Approved | + ✎ |
| > Developmental stage | Keywords assigned to proteins because they are expressed specifically in a given developmental stage. | No Owner | Approved | + ✎ |
| > Disease | Keywords assigned to proteins because they are involved in a specific disease. | No Owner | Approved | + ✎ |
| > Domain | Keywords assigned to proteins because they have at least one specimen of a specific domain. | No Owner | Approved | + ✎ |
| > Ligand | Keywords assigned to proteins because they bind, are associated with, or whose activity is dependent of some molecule. | No Owner | Approved | + ✎ |
| > Molecular function | Keywords assigned to proteins due to their particular molecular function. | No Owner | Approved | + ✎ |
| > PTM | Keywords assigned to proteins because their sequences can differ from the mere translation of their corresponding... more | No Owner | Approved | + ✎ |
| > Technical term | Keywords assigned to proteins according to technical reasons. | No Owner | Approved | + ✎ |

B

```
"parent","name*","displayName","description","synonyms","relatedTerms","references","tags","reviewers","owner","status"
"","Technical term","Technical term","Keywords assigned to proteins according to 'technical' reasons.","","","","","",""
"","PTM","PTM","Keywords assigned to proteins because their sequences can differ from the mere translation of their corresponding genes, due to some post- translational modification.","","","","","",""
"","Molecular function","Molecular function","Keywords assigned to proteins due to their particular molecular function.","","","","","",""
"","Ligand","Ligand","Keywords assigned to proteins because they bind, are associated with, or whose activity is dependent of some molecule.","","","","","",""
"","Domain","Domain","Keywords assigned to proteins because they have at least one specimen of a specific domain.","","","","","",""
"","Disease","Disease","Keywords assigned to proteins because they are involved in a specific disease.","","","","","",""
"","Developmental stage","Developmental stage","Keywords assigned to proteins because they are expressed specifically in a given developmental stage.","","","","","",""
"","Coding sequence diversity","Coding sequence diversity","Keywords assigned to proteins because their sequences can differ, due to differences in the coding sequences such as polymorphisms, RNA- editing, alternative splicing.","","","","","",""
"","Cellular component","Cellular component","Keywords assigned to proteins because they are found in a specific cellular or extracellular component.","","","","","",""
"","Biological process","Biological process","Keywords assigned to proteins because they are involved in a particular biological process.","","","","","",""
"Interactome.Ligand","2Fe-2S","2Fe-2S","Protein which contains at least one 2Fe-2S iron-sulfur cluster: 2 iron atoms complexed to 2 inorganic sulfides and 4 sulfur atoms of cysteines from the protein.","[2Fe-2S] cluster; [Fe2S2] cluster; 2 iron; 2 sulfur cluster binding; Di-mu-sulfido-diiron; Fe2/S2 (inorganic) cluster; Fe2S2","","","","",""
"Interactome.Technical term","3D-structure","3D-structure","Protein, or part of a protein, whose three-dimensional structure has been resolved experimentally (for example by X-ray crystallography or NMR spectroscopy) and whose coordinates are available in the PDB database. Can also be used for theoretical models.","","","","","",""
"Interactome.Ligand","3Fe-4S","3Fe-4S","Protein which contains at least one 3Fe-4S iron-sulfur cluster: 3 iron atoms complexed to 4 inorganic sulfides and 3 sulfur atoms of cysteines from the protein. In a number of iron-sulfur proteins, the 4Fe-4S cluster can be reversibly converted by oxidation and loss of one iron ion to a 3Fe-4S cluster.","","","","","",""
"Interactome.Ligand","4Fe-4S","4Fe-4S","Protein which contains at least one 4Fe-4S iron-sulfur cluster: 4 iron atoms complexed to 4 inorganic sulfides and 4 sulfur atoms of cysteines from the protein. In a number of iron-sulfur proteins, the 4Fe-4S cluster can be reversibly converted by oxidation and loss of one iron ion to a 3Fe-4S cluster.","","","","","",""
"Interactome.Biological process","Acetoin biosynthesis","Acetoin biosynthesis","Protein involved in the synthesis of acetoin (3-hydroxy-2-butanone). Acetoin is a component of the butanediol cycle (butanediol fermentation) in microorganisms.","3-hydroxy-2-butanone anabolism; 3-hydroxy-2-butanone biosynthesis; 3-hydroxy-2-butanone biosynthetic process; 3-hydroxy-2-butanone formation; 3-hydroxy-2-butanone synthesis; Acetoin anabolism; Acetoin biosynthetic process; Acetoin formation; Acetoin synthesis","","","","",""
"Interactome.Biological process","Acetoin catabolism","Acetoin catabolism","Protein involved in the degradation of acetoin (3-hydroxy-2-butanone). Acetoin is a component of the butanediol cycle (butanediol fermentation) in microorganisms.","3-hydroxy-2-butanone breakdown; 3-hydroxy-2-butanone catabolic process; 3-hydroxy-2-butanone degradation; Acetoin breakdown; Acetoin catabolic process; Acetoin degradation","","","","",""
"Interactome.PTM","Acetylation","Acetylation","Protein which is posttranslationally modified by the attachment of at least one acetyl group; generally at the N-terminus.","Acetylated; N-acetylated","","","","",""
"Interactome.Molecular function","Acetylcholine receptor inhibiting toxin","Acetylcholine receptor inhibiting toxin","Toxin which interferes with the function of the nicotinic acetylcholine receptor (nAChR). The nAChR is a postsynaptic membrane protein that undergoes an extensive conformational change upon binding to acetylcholine, leading to opening of an ion-conducting channel across the plasma membrane. These toxins are mostly found in snake and cone snail venoms.","nAChR inhibitor; Nicotinic AChR inhibitor","","","","",""
"Interactome.Molecular function","Actin-binding","Actin-binding","Protein which binds to actin, and thereby can modulate the properties and/or functions of the actin filament.","Actin filament binding","","","","",""
"Interactome.Molecular function","Activator","Activator","Protein that positively regulates either the transcription of one or more genes, or the translation of mRNA.","Positive activator","","","","",""
```

**Figure 6: Preparing UniProt glossary data for import into OpenMetadata.** a) The UniProt glossary contains the parent terms and sub-terms together. OpenMetadata requires the sub-term records processed with glossaryname.parent for parent column and parent term records with empty string. This requires the same file be processed differently and then recombined back together. Here we show the final processing steps with the parent records DataFrame already stored out in a prior stage. In stage 19, a literal value is created and followed by a stage to combine with parent name and rename the column back to parent. A union stage combines the stored DataFrame containing the parent term records with modified sub-term DataFrame. Finally, a select is done to ensure correct column ordering before writing out as a CSV file. b) Result of the UniProt glossary data loaded into OpenMetaData and named "Interactome". The parent terms are shows with a symbol to expand to sub-terms

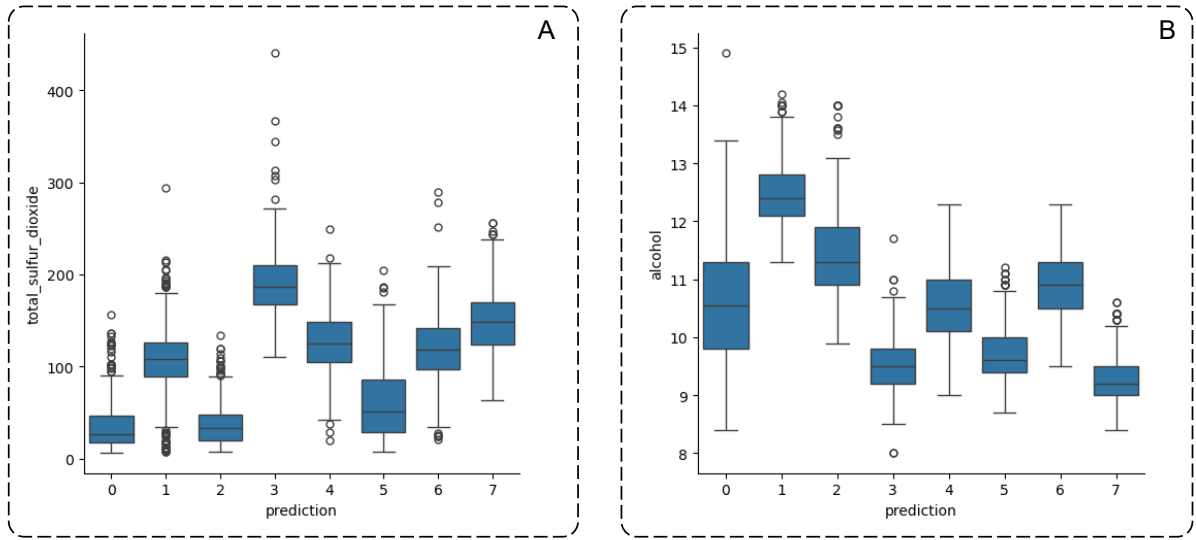

**Figure 7: Plot of total sulfur dioxide vs predicted cluster for wine quality data.** This plot was generated using the edaBoxplot stage with parameter selection set for processing of all combinations of numerical and categorical features. Out of all generated plots, this one had a visible separation of values for each cluster.